\documentclass[11pt]{article}

\usepackage[preprint]{acl}

\usepackage{times}
\usepackage{latexsym}
\usepackage[T1]{fontenc}
\usepackage[utf8]{inputenc}
\usepackage{microtype}
\usepackage{inconsolata}
\usepackage{graphicx}
\usepackage{booktabs}
\usepackage{multirow}
\usepackage{amsmath}
\usepackage{xcolor}

\newcommand{\jev}{Jev}
\newcommand{\luna}{GPT-6 Luna}

\usepackage[most]{tcolorbox}
\usepackage{enumitem}
\usepackage{iftex}

\ifPDFTeX
  \usepackage{CJKutf8}
  \newcommand{\zh}[1]{\begin{CJK}{UTF8}{gbsn}#1\end{CJK}}
\else
  \usepackage{xeCJK}
  \newcommand{\zh}[1]{#1}
\fi

\definecolor{promptaccent}{HTML}{2F5D8A}
\newcommand{\pfield}[1]{\textbf{#1.}\enspace}

\newtcolorbox{promptbox}[2][]{%
  enhanced, breakable,
  title={#2},
  fonttitle=\normalsize\bfseries, coltitle=black,
  colbacktitle=black!8,
  colback=white, colframe=black,
  boxrule=0.5pt, arc=1pt,
  left=10pt, right=10pt, top=6pt, bottom=6pt,
  toptitle=3pt, bottomtitle=3pt, lefttitle=10pt,
  titlerule=0.4pt,
  fontupper=\small,
  before skip=10pt, after skip=14pt,
  before upper={\setlength{\parskip}{3pt}%
    \setlist[itemize]{leftmargin=1.6em, itemsep=3pt,
                      topsep=3pt, parsep=0pt}},
  break at=-\baselineskip/0pt,
  #1}

\title{Decision Models for Financial NLP: A Benchmark of Jev and OpenAI's Decisions API}

\author{
  Xingyuan Liu \\
  HKU \\
  \texttt{xingyuan.liu@connect.hku.hk}
  \And
  Allen H. Huang \\
  HKUST \\
  \texttt{allen.huang@ust.hk}
  \And
  Yi Yang \\
  HKUST \\
  \texttt{imyiyang@ust.hk}
}

\begin{document}
\maketitle

\begin{abstract}
Decision models have recently emerged as a new approach to structured probabilistic prediction, with systems such as Jev and OpenAI's Decisions API attracting growing interest. We present a systematic benchmark of two emerging decision models, Jev and GPT-6 Luna, across 21 financial tasks, covering text classification, text-pair classification, earnings-surprise prediction, and stock-movement prediction. Our results show that decision models outperform embedding-based classifiers on 12 of 13 single-text classification tasks, although task-specific fine-tuned encoders remain stronger on their specialized tasks. However, their performance on financial outcome prediction tasks shows limited improvements over simple baselines. Together with their low inference costs, these findings suggest that decision models offer a promising approach to zero-shot financial text classification, although they do not yet match the performance of task-specific fine-tuned encoders.
\end{abstract}

\section{Introduction}
Financial text analysis plays an important role in financial research and practice, supporting a wide range of applications such as sentiment classification, corporate disclosure analysis, and financial forecasting. Over the past decade, natural language processing (NLP) methods have evolved from dictionary-based approaches to domain-specific pretrained language models and large language models (LLMs) \citep{loughran2011liability,huang2023finbert,wu2023bloomberggpt}. In particular, encoder-based models such as FinBERT have become widely used for extracting financial information from unstructured text, while recent LLMs have demonstrated the potential to perform financial NLP tasks through natural-language instructions without task-specific fine-tuning \citep{guo2023chatgpt,xie2023pixiu}. These advances have motivated growing interest in evaluating the capabilities of emerging AI models across financial applications.

Recently, a new class of AI models, referred to as \textit{System One decision models}, has emerged as an alternative approach to structured prediction and classification. Unlike conventional generative LLM interfaces, decision models are designed to answer explicitly defined questions by returning probabilities over predefined options. For example, given a financial news sentence stating that ``the company reported a 20\% decline in quarterly earnings," a decision model can directly assign probabilities to three sentiment categories: positive (0.05), neutral (0.10), and negative (0.85), rather than generating a free-form textual response. TypeSafe's Jev enables such probabilistic judgments through natural-language decision criteria without task-specific fine-tuning \citep{typesafe2026jev}. OpenAI has also introduced a Decisions API that supports structured probabilistic decisions \citep{openai2026decisions}. 

Despite these developments, the effectiveness of decision models in financial NLP remains largely unexplored. Financial applications encompass diverse tasks, ranging from sentiment analysis and disclosure classification to semantic matching and predictions of future financial outcomes. Existing financial NLP benchmarks have primarily focused on encoder-based models, embedding models, and generative LLMs \citep{li2023chatgpt,xie2023pixiu,tang2025finmteb}. However, little is known about how emerging decision models perform on these tasks or how they compare with established approaches. A systematic evaluation is therefore needed to assess their capabilities, identify their strengths and limitations, and provide empirical evidence for researchers and practitioners considering their use in financial applications.

In this paper, we benchmark two emerging decision models, TypeSafe's Jev and GPT-6 Luna through OpenAI's Decisions API, across 21 diverse financial NLP tasks. Our evaluation covers 13 single-text classification tasks, four text-pair classification tasks, and four financial prediction tasks. The classification tasks, drawn from FinMTEB \citep{tang2025finmteb}, span English and Chinese financial text and include sentiment analysis, forward-looking statement identification, ESG classification, monetary policy stance, and financial news analysis. The prediction tasks examine next-quarter earnings surprises using earnings-call transcripts \citep{fincall2026} and stock-price movements using financial social media data \citep{xu2018stocknet,wu2018hybrid,soun2022accurate}. We evaluate both decision models in a zero-shot setting and compare their performance with established fine-tuned financial encoders, embedding-based classifiers, and available LLM and financial prediction baselines.

Our evaluation yields three main findings. First, task-specific fine-tuned encoders consistently outperform decision models on the classification tasks for which they were trained, highlighting the continued value of task-specific supervision. Second, zero-shot decision models demonstrate competitive performance on financial text classification, outperforming embedding-based classifiers without task-specific fine-tuning on 12 of 13 single-text classification tasks. Third, decision models show limited effectiveness on financial outcome prediction. On earnings-surprise prediction, they achieve only modest discriminative performance, comparable to a simple FinBERT-based sentiment baseline, while their stock-movement predictions remain close to chance across three datasets. Overall, these findings suggest that decision models offer a promising zero-shot approach to financial text classification, but do not yet  demonstrate reliable advantages in financial outcome prediction. We hope this benchmark serves as a useful reference for researchers and practitioners evaluating emerging decision models for financial applications.

\section{Related Work}

\paragraph{Models and benchmarks for financial NLP.}
Financial NLP has evolved from dictionary-based methods
\citep{loughran2011liability} to domain-specific pretrained
language models and general-purpose LLMs. FinBERT
\citep{yang2020finbert,huang2023finbert} was pretrained on
financial communications and subsequently fine-tuned for
tasks such as financial sentiment, forward-looking statement
identification, and ESG classification. More recent studies
have evaluated the capabilities of generative LLMs for
financial text analysis, finding that their zero-shot
performance varies across tasks and that fine-tuned models
remain competitive on specialized classification tasks
\citep{li2023chatgpt,guo2023chatgpt,xie2023pixiu}.
FinMTEB \citep{tang2025finmteb} further provides a
systematic evaluation of embedding models on financial
NLP tasks using supervised classifiers trained on
text embeddings.

\paragraph{Financial text for financial outcome prediction.}
Beyond NLP tasks, financial text has been used to
predict future financial outcomes. StockNet
\citep{xu2018stocknet} investigates stock movement
prediction using social media text and historical prices,
while subsequent studies introduce additional benchmarks
and prediction models based on financial social media
\citep{wu2018hybrid,soun2022accurate}.
More recently, FinCall-Surprise \citep{fincall2026}
evaluates whether earnings-call transcripts can predict
subsequent earnings surprises, reporting limited
predictive performance from prompted LLMs.
These studies provide established tasks and reference
results for evaluating financial prediction models.

\paragraph{Emerging decision models.}
Decision models have recently emerged as a new approach
to structured probabilistic prediction. Unlike conventional
generative LLM interfaces, they directly return
probabilities over predefined decision options.
TypeSafe's \jev{} supports multiple output formats,
including Choice, Score, and Noul, and is described
as being trained through reinforcement learning for
calibrated decisions \citep{typesafe2026jev}.
OpenAI has also introduced a Decisions API for structured
probabilistic decisions \citep{openai2026decisions}.
An early evaluation by \citet{denney2026jev} compares
decision models with conventional LLMs on political
science classification tasks, finding broadly comparable
performance, while specialized fine-tuned encoders
remain competitive \citep{brandt2025extractive}.
However, decision models have not yet been systematically
benchmarked across financial NLP tasks. Our study
addresses this gap by evaluating emerging decision models
on financial text classification, text-pair classification,
and financial prediction tasks.
\section{Methods}

We benchmark emerging decision models against established
financial NLP approaches across financial NLP tasks, and financial outcome prediction tasks. All decision models are evaluated in a zero-shot setting,
without task-specific fine-tuning.

\subsection{Models and Baselines}
\paragraph{\jev{}.}
We evaluate TypeSafe's \jev{} \citep{typesafe2026jev},
using version \texttt{jev-1.13.0} with the Choice
output format. The model receives a text and a
structured question with predefined answer options,
and returns a probability for each option.

\paragraph{\luna{}.}
We evaluate \luna{} through OpenAI's Decisions API
\citep{openai2026decisions}, using
\texttt{gpt-6-luna} via \texttt{POST /v1/decisions}.
We use the same zero-shot prompts and answer
options as \jev{}.

\paragraph{Fine-tuned encoders.}
We evaluate three publicly available task-specific checkpoints: \texttt{yiyanghkust/finbert-tone}, \texttt{finbert-fls}, and \texttt{finbert-esg} \citep{huang2023finbert}. Each checkpoint is evaluated only on the task for which it was fine-tuned: financial sentiment (FinSent), forward-looking statements (FLS), and ESG classification, respectively. All checkpoints run locally using Apple MPS (batch size 32, maximum sequence length 512 tokens). 

\paragraph{Embedding-based classifiers.}
For classification and text-pair classification, we use results from the FinMTEB leaderboard \citep{tang2025finmteb}. For classification, FinMTEB fits a few-shot logistic regression classifier on frozen embeddings; for text-pair classification, it scores pairs by embedding similarity without training. For English tasks, we report BERT and FinBERT [CLS] embeddings, along with the best remaining embedding model for each task. For Chinese tasks, we report bag-of-words (BOW), multilingual MiniLM, and the best remaining embedding model.

\paragraph{Financial outcome prediction baselines.}
For earnings-surprise prediction, we include generative LLM results reported by \citet{fincall2026}, a majority-class baseline, and a FinBERT sentiment baseline. For the latter, we score each sentence in the earnings-call transcript using \texttt{finbert-tone} and average $P(\mathrm{positive})-P(\mathrm{negative})$ across sentences. For stock-movement prediction, we use the majority-class baseline and the same FinBERT sentiment approach, averaging sentiment scores across the available tweets. The sign of the average score determines the predicted direction.

\subsection{Evaluation Protocol}

\paragraph{Zero-shot decision queries.}
For each task, we construct one structured decision question with natural-language instructions and predefined answer options. Each option includes a criterion describing its corresponding category. Both decision models receive the same instructions and option definitions, without labeled examples or task-specific fine-tuning. All prompts are provided in Appendix~\ref{app:prompts}.

\paragraph{Input processing.}
For single-text classification, each input contains one financial text sample. For text-pair classification, both texts are provided in a single query, asking whether they match on the specified dimension.
For earnings-surprise prediction, the input is an earnings-call transcript. Transcripts longer than 31K tokens are shortened for \jev{} and \luna{} using the procedure described in Appendix~\ref{app:prompts}.
For stock-movement prediction, tweets from the five days preceding the target trading day are grouped by trading day and provided without historical price information.

\paragraph{Evaluation metrics.}
For single-text classification, we report accuracy. For text-pair classification, we report average precision (AP), following FinMTEB \citep{tang2025finmteb}, using the probability of the ``match'' option as the ranking score.
For earnings-surprise prediction, we report accuracy, macro-averaged precision, recall, and F1 following \citet{fincall2026}, together with Matthews correlation coefficient (MCC) and AUROC. AUROC is computed using the probability of an earnings beat for decision models and the mean sentiment score for FinBERT.
For stock-movement prediction, we report accuracy, MCC, and AUROC, using the probability of ``up'' as the ranking score for decision models and the mean sentiment score for FinBERT.

\section{Tasks}

We evaluate decision models on 21 financial NLP tasks,
covering text classification, text-pair classification,
and financial prediction (Table~\ref{tab:tasks}).
We group these tasks into measurement tasks, which
identify information expressed in financial text,
and prediction tasks, which forecast future financial
outcomes.

\subsection{Financial NLP Tasks}

We use 17 tasks from FinMTEB \citep{tang2025finmteb},
including 13 single-text classification tasks and
four text-pair classification tasks in English and
Chinese.

\paragraph{English sentiment classification.}
Financial PhraseBank \citep{malo2014good} contains
financial news sentences labeled for investor
sentiment. FinSent \citep{huang2023finbert}
contains analyst-report sentences labeled for
financial sentiment. SemEval-17
\citep{cortis2017semeval} evaluates sentiment
toward financial entities in news headlines,
while FiQA-ABSA \citep{maia2018fiqa} focuses on
aspect-level financial sentiment.

\paragraph{Financial disclosure and policy classification.}
FLS \citep{huang2023finbert} classifies sentences
from 10-K filings as non-forward-looking,
non-specific forward-looking, or specific
forward-looking statements.
ESG \citep{huang2023finbert} identifies
environmental, social, governance, or
non-ESG content.
FOMC \citep{shah2023trillion} classifies
monetary policy statements as hawkish,
dovish, or neutral.
FinancialFraud \citep{tang2025finmteb}
identifies evidence of financial fraud
in corporate filings. For long documents,
we use deterministic excerpts of up to
16,000 characters.

\paragraph{Chinese financial classification.}
FinNSP and FinFE \citep{lu2023bbtfin}
evaluate negative financial news detection
and social-media sentiment, respectively.
FinChinaSentiment \citep{lan2023chinese}
classifies the severity of financial risk
warnings. OpenFinData \citep{tang2025finmteb}
evaluates financial sentiment, while Weibo21
\citep{nan2021mdfend} identifies fake
financial news.

\paragraph{Text-pair classification.}
We evaluate three financial headline matching
tasks: HeadlineAC, HeadlinePDD, and HeadlinePDU
\citep{sinha2021impact,tang2025finmteb}.
We also include AFQMC, a Chinese financial
question-matching task \citep{tang2025finmteb}.
These tasks evaluate whether two texts
match according to a specified semantic
relationship.

\subsection{Financial Outcome Prediction Tasks}

We evaluate four financial prediction tasks:
one earnings-surprise prediction task and three
stock-movement prediction tasks.

\paragraph{Earnings-surprise prediction.}
FinCall-Surprise \citep{fincall2026} contains
earnings-call transcripts from 2019--2021,
with labels indicating whether a company's
next-quarter earnings per share (EPS) beats
or misses analyst consensus estimates.
We use the transcript text to predict
the direction of the subsequent earnings
surprise.

\paragraph{Stock-movement prediction.}
We use StockNet \citep{xu2018stocknet},
CIKM18 \citep{wu2018hybrid}, and BigData22
\citep{soun2022accurate}, which contain
financial social-media posts and stock-movement
labels. Each sample includes tweets from
the five days preceding the target trading day.
Following the StockNet setting, we exclude
movements between -0.5\% and 0.55\%.
We evaluate models in a text-only setting,
without historical stock prices.
For StockNet, we use the original test period
(October--December 2015). For CIKM18 and
BigData22, we use the last three months
as the test period.

\begin{table}[t]
\centering
\small
\setlength{\tabcolsep}{3pt}
\begin{tabular}{@{}llrl@{}}
\toprule
Task & Content & $n_{\text{test}}$ & $K$ \\
\midrule
PhraseBank & investor sentiment, news & 1000 & 3 \\
FinSent & analyst-report sentiment & 1000 & 3 \\
SemEval-17 & headline sentiment & 343 & 3 \\
FiQA-ABSA & aspect sentiment & 352 & 2 \\
FLS & forward-looking (10-K) & 1000 & 3 \\
ESG & ESG topic & 1000 & 4 \\
FOMC & hawkish/dovish stance & 1000 & 3 \\
FinancialFraud & fraud evidence (excerpt) & 51 & 2 \\
\midrule
FinNSP\textsuperscript{zh} & negative news & 1000 & 2 \\
FinChinaSent.\textsuperscript{zh} & warning severity & 1000 & 4 \\
FinFE\textsuperscript{zh} & social-media sentiment & 1000 & 3 \\
OpenFinData\textsuperscript{zh} & sentiment & 23 & 3 \\
Weibo21\textsuperscript{zh} & fake financial news & 238 & 2 \\
\midrule
HeadlineAC & headline pair matching & 7285 & 2 \\
HeadlinePDD & headline pair matching & 9197 & 2 \\
HeadlinePDU & headline pair matching & 9697 & 2 \\
AFQMC\textsuperscript{zh} & question pair matching & 34334 & 2 \\
\midrule
FinCall-Surprise & call $\rightarrow$ EPS surprise & 2688 & 2 \\
StockNet & tweets $\rightarrow$ movement & 2632 & 2 \\
CIKM18 & tweets $\rightarrow$ movement & 1231 & 2 \\
BigData22 & tweets $\rightarrow$ movement & 2334 & 2 \\
\bottomrule
\end{tabular}
\caption{Tasks used in this study. $K$ is the number of class labels; zh = Chinese.}
\label{tab:tasks}
\end{table}

\section{Experiments}
\subsection{Financial NLP Tasks}
\label{sec:financial_nlp}

Tables~\ref{tab:finetuned} and~\ref{tab:nonfinetuned} report single-text classification results, while Table~\ref{tab:pair} reports text-pair classification results.

\paragraph{Fine-tuned encoders remain stronger on their specialized tasks.}
As shown in Table~\ref{tab:finetuned}, task-specific fine-tuned encoders outperform both decision models on all three tasks for which they were trained, with accuracy gaps ranging from 5.2 to 14.2 percentage points. The smallest gap occurs on ESG classification, while the largest occurs on FLS classification. These results highlight the continued advantage of task-specific fine-tuning for established financial classification tasks.

\paragraph{Decision models perform strongly on zero-shot financial text classification.}
Table~\ref{tab:nonfinetuned} shows that either \jev{} or \luna{} achieves the highest accuracy on 12 of 13 single-text classification tasks, outperforming embedding-based classifiers even though the latter fit logistic regression probes on eight labeled examples per class. The improvements are particularly notable on FLS, FinFE, and OpenFinData. FinancialFraud is the only exception, where the best embedding-based classifier achieves 0.77 accuracy, compared with 0.65 for both decision models.

\paragraph{Performance is mixed on text-pair classification.}
As shown in Table~\ref{tab:pair}, decision models achieve the highest average precision on two of four tasks (HeadlineAC and AFQMC), but fall behind embedding similarity on HeadlinePDD and HeadlinePDU. Their strong performance on single-text classification therefore does not consistently extend to text-pair classification.
\begin{table}[t]
\centering
\small
\setlength{\tabcolsep}{3.5pt}
\resizebox{\columnwidth}{!}{%
\begin{tabular}{@{}lccccc@{}}
\toprule
Task & BOW & FinBERT [CLS] & Best other & \jev{} & \luna{} \\
\midrule
HeadlineAC  & 0.77 & 0.75 & 0.78$^{a}$ & \textbf{0.89} & 0.83 \\
HeadlinePDD & 0.60 & 0.59 & \textbf{0.85}$^{a}$ & 0.72 & 0.72 \\
HeadlinePDU & 0.77 & 0.75 & \textbf{0.78}$^{a}$ & 0.75 & 0.72 \\
AFQMC\textsuperscript{zh} & 0.34 & -- & 0.60$^{b}$ & \textbf{0.63} & 0.62 \\
\bottomrule
\end{tabular}%
}
\caption{Pair-classification average precision (AP).
Encoder values are from the FinMTEB leaderboard;
$^{a}$Fin-e5, $^{b}$gte-Qwen1.5-7B-instruct.}
\label{tab:pair}
\end{table}

\begin{table}[t]
\centering
\small
\setlength{\tabcolsep}{0pt} 
\begin{tabular*}{\columnwidth}{@{\extracolsep{\fill}}llccc@{}}
\toprule
Task & Checkpoint & Fine-tuned & \jev{} & \luna{} \\
\midrule
FinSent & finbert-tone & \textbf{0.93} & 0.79 & 0.81 \\
FLS & finbert-fls & \textbf{0.92} & 0.80 & 0.78 \\
ESG & finbert-esg & \textbf{0.99} & 0.94 & 0.94 \\
\bottomrule
\end{tabular*}
\caption{Accuracy of fine-tuned checkpoints and zero-shot decision models on the tasks the checkpoints were trained for.}
\label{tab:finetuned}
\end{table}

\begin{table*}[t]
\centering
\small
\setlength{\tabcolsep}{5pt}
\begin{tabular}{@{}l ccc cc@{}}
\toprule
& \multicolumn{3}{c}{FinMTEB}
& \multicolumn{2}{c}{Zero-shot} \\
\cmidrule(lr){2-4}\cmidrule(lr){5-6}
English & BERT [CLS] & FinBERT [CLS] & Best other
        & \jev{} & \luna{} \\
\midrule
PhraseBank     & 0.67 & 0.73 & 0.86$^{a}$ & 0.92 & \textbf{0.96} \\
FinSent        & 0.47 & 0.53 & 0.72$^{a}$ & 0.79 & \textbf{0.81} \\
FiQA-ABSA      & 0.56 & 0.59 & 0.90$^{a}$ & \textbf{0.92} & 0.89 \\
SemEval-17     & 0.46 & 0.50 & 0.74$^{a}$ & \textbf{0.77} & 0.65 \\
FLS            & 0.56 & 0.57 & 0.61$^{b}$ & \textbf{0.80} & 0.78 \\
ESG            & 0.69 & 0.77 & 0.91$^{a}$ & \textbf{0.94} & \textbf{0.94} \\
FOMC           & 0.37 & 0.39 & 0.65$^{a}$ & \textbf{0.73} & 0.66 \\
FinancialFraud & 0.62 & 0.66 & \textbf{0.77}$^{c}$ & 0.65 & 0.65 \\
\midrule
Chinese & BOW & mMiniLM & Best other
        & \jev{} & \luna{} \\
\midrule
FinNSP         & 0.71 & 0.80 & 0.93$^{d}$ & \textbf{0.96} & 0.95 \\
FinChinaSent.  & 0.29 & 0.32 & 0.35$^{e}$ & 0.39 & \textbf{0.40} \\
FinFE          & 0.37 & 0.46 & 0.56$^{f}$ & \textbf{0.71} & \textbf{0.71} \\
OpenFinData    & 0.34 & 0.55 & 0.64$^{e}$ & 0.87 & \textbf{0.91} \\
Weibo21        & 0.64 & 0.67 & 0.74$^{f}$ & \textbf{0.86} & 0.81 \\
\bottomrule
\end{tabular}

\caption{Classification accuracy of zero-shot decision models and encoders without task-specific fine-tuning. Encoder scores are FinMTEB leaderboard values. For Chinese, where no BERT or FinBERT scores are available, we report a bag-of-words baseline (BOW) and paraphrase-multilingual-MiniLM-L12-v2 (mMiniLM). ``Best other'' is the best remaining leaderboard model per task: $^{a}$Fin-e5, $^{b}$voyage-3-large, $^{c}$e5-mistral-7b-instruct, $^{d}$text-embedding-3-large, $^{e}$bge-m3, $^{f}$gte-Qwen1.5-7B-instruct (tied with bge-large-zh-v1.5 on FinFE).}
\label{tab:nonfinetuned}
\end{table*}

\begin{table}[t]
\centering
\small
\setlength{\tabcolsep}{3pt}
\resizebox{\columnwidth}{!}{%
\begin{tabular}{@{}lcccccc@{}}
\toprule
Model & Acc & M-P & M-R & M-F1 & MCC & AUROC \\
\midrule
Majority (beat) & 0.85 & 0.43 & 0.50 & 0.46 & 0.00 & 0.500 \\
\midrule
FinBERT tone sign & 0.84 & 0.49 & 0.50 & 0.47 & $-$0.01 & 0.561 \\
\midrule
\jev{}  & 0.67 & \textbf{0.53} & \textbf{0.55} & 0.52 & \textbf{0.07} & 0.563 \\
\luna{} & 0.62 & 0.52 & 0.54 & 0.50 & 0.06 & \textbf{0.564} \\
\midrule
GPT-oss 20B$^{\dagger}$     & 0.71 & \textbf{0.53} & 0.54 & \textbf{0.53} & -- & -- \\
Qwen-2.5 14B$^{\dagger}$    & \textbf{0.80} & 0.51 & 0.50 & 0.50 & -- & -- \\
Mistral 7B$^{\dagger}$      & 0.74 & 0.51 & 0.51 & 0.51 & -- & -- \\
Llama-3.2 3B$^{\dagger}$    & 0.78 & 0.51 & 0.51 & 0.51 & -- & -- \\
Gemma-3 1B$^{\dagger}$      & 0.74 & 0.44 & 0.45 & 0.45 & -- & -- \\
Fin-LLaMA 8B$^{\dagger}$    & 0.74 & 0.47 & 0.47 & 0.47 & -- & -- \\
Finance-R1 1.7B$^{\dagger}$ & 0.52 & 0.40 & 0.33 & 0.36 & -- & -- \\
\bottomrule
\end{tabular}%
}
\caption{FinCall-Surprise results: accuracy and macro-averaged precision, recall and F1, MCC and AUROC. The three AUROC values do not differ significantly. $^{\dagger}$Reported by \citet{fincall2026}, who do not report MCC or AUROC. }
\label{tab:fincall}
\end{table}

\begin{table}[t]
\centering
\small
\setlength{\tabcolsep}{3pt}
\resizebox{\columnwidth}{!}{%
\begin{tabular}{@{}lcccc@{}}
\toprule
Model & Acc & MCC & AUROC & Up \\
\midrule
\multicolumn{5}{@{}l}{\emph{StockNet} } \\
Majority (up)     & 0.53 & 0.00 & 0.50 & 1.00 \\
FinBERT tone sign & 0.51 & $-$0.01 & 0.49 & 0.69 \\
\jev{}            & 0.52 & 0.02 & 0.51 & 0.62 \\
\luna{}           & 0.51 & $-$0.02 & 0.50 & 0.83 \\
\midrule
\multicolumn{5}{@{}l}{\emph{CIKM18}} \\
Majority (up)     & 0.55 & 0.00 & 0.50 & 1.00 \\
FinBERT tone sign & 0.54 & 0.00 & 0.53 & 0.91 \\
\jev{}            & 0.54 & 0.06 & 0.53 & 0.65 \\
\luna{}           & 0.55 & 0.05 & 0.52 & 0.73 \\
\midrule
\multicolumn{5}{@{}l}{\emph{BigData22}} \\
Majority (up)     & 0.53 & 0.00 & 0.50 & 1.00 \\
FinBERT tone sign & 0.53 & 0.03 & 0.50 & 0.79 \\
\jev{}            & 0.50 & $-$0.02 & 0.49 & 0.61 \\
\luna{}           & 0.51 & 0.00 & 0.50 & 0.77 \\
\bottomrule
\end{tabular}%
}
\caption{Next-day stock movement from tweets (text only). Up = share of samples predicted up. }
\label{tab:movement}
\end{table}

\subsection{Financial Outcome Prediction Tasks}
\label{sec:financial_prediction}

Tables~\ref{tab:fincall} and~\ref{tab:movement} report earnings-surprise and stock-movement prediction results, respectively.

\paragraph{Earnings-surprise prediction.}
On FinCall-Surprise (Table~\ref{tab:fincall}), \jev{} and \luna{} achieve AUROCs of 0.563 and 0.564, respectively, comparable to the FinBERT sentiment baseline (0.561). Their macro-F1 scores (0.52 and 0.50) are also comparable to those of the generative LLMs reported by \citet{fincall2026}. Although both decision models identify some earnings misses, their overall discriminative performance remains modest, with no clear advantage over the FinBERT baseline.

\paragraph{Stock-movement prediction.}
On StockNet, CIKM18, and BigData22 (Table~\ref{tab:movement}), both decision models achieve AUROCs between 0.49 and 0.53, with MCC values close to zero. Their performance is comparable to the FinBERT sentiment baseline and does not consistently exceed majority-class accuracy. These results suggest limited predictive value for next-day stock movements when only financial social media text is available.

\section{Further Analyses}
\label{sec:further}

\begin{table}[t]
\centering
\small
\setlength{\tabcolsep}{4pt}
\begin{tabular}{@{}llccc@{}}
\toprule
Task & Class & Fine-tuned & \jev{} & \luna{} \\
\midrule
\multirow{3}{*}{FLS}
 & Not FLS          & \textbf{0.97} & 0.86 & 0.88 \\
 & Non-specific FLS & \textbf{0.91} & 0.65 & 0.53 \\
 & Specific FLS     & 0.81 & 0.86 & \textbf{0.89} \\
\midrule
\multirow{4}{*}{ESG}
 & Environmental & \textbf{0.99} & 0.95 & 0.96 \\
 & Social        & \textbf{0.99} & 0.85 & 0.89 \\
 & Governance    & \textbf{1.00} & 0.92 & 0.78 \\
 & Non-ESG       & \textbf{1.00} & 0.99 & 0.98 \\
\bottomrule
\end{tabular}
\caption{Per-class recall on FLS and ESG. Fine-tuned = \texttt{finbert-fls} and \texttt{finbert-esg}.}
\label{tab:perclass}
\end{table}

\paragraph{Per-class errors.} Table~\ref{tab:perclass} reports per-class recall for FLS and ESG classification. Decision models perform relatively well on clearly defined categories, such as Environmental, Non-ESG, and Specific FLS, but show larger gaps relative to fine-tuned encoders on Non-specific FLS, Social, and Governance. For example, on Non-specific FLS, \jev{} and \luna{} achieve recall of 0.65 and 0.53, respectively, compared with 0.91 for the fine-tuned encoder. These results suggest that task-specific fine-tuning remains valuable for distinguishing closely related categories.

\paragraph{Cost and speed.}
Decision models offer low API inference costs without requiring task-specific fine-tuning. \jev{} bills input tokens only. In our experiments, this translates to \$0.021--0.033 per 1{,}000 sentence-level decisions. \luna{} charges \$0.10 per million uncached input tokens with no output-token charges, corresponding to \$0.022--0.053 per 1{,}000 English sentence-level decisions (\$0.12 for longer Chinese risk warnings). Median latency is approximately 0.8\,s per sequential request for \jev{} and 0.6--0.7\,s for \luna{}. Although fine-tuned encoders generally offer inexpensive local inference, they require task-specific labeled data and training. Decision models therefore provide a convenient alternative when such resources are unavailable.
\section{Conclusion}

In this paper, we benchmarked two emerging decision models, \jev{} and \luna{}, across 21 financial tasks. Our results show that decision models achieve strong zero-shot performance on financial text classification, outperforming embedding-based classifiers on 12 of 13 single-text classification tasks, while task-specific fine-tuned encoders remain stronger on their specialized tasks. Their advantages are less consistent on text-pair classification, and their performance on financial outcome prediction remains limited. Importantly, decision models offer these classification capabilities at low inference costs without task-specific fine-tuning, making them a promising alternative for financial text analysis. We hope this benchmark serves as a useful reference for researchers and practitioners evaluating emerging decision models for financial applications.

\section*{Limitations}

This study has several limitations. First, the decision models may have encountered public benchmark datasets during pretraining, raising concerns about data contamination \citep{sainz2023nlp}. For financial outcome prediction, there is also a risk of look-ahead bias, as the models may have been exposed to information released after the historical prediction dates. Although our prompts restrict the models to the provided text, we cannot rule out the influence of future information embedded in their pretrained parameters. Second, the baseline results come from different evaluation settings. FinMTEB embedding models use supervised logistic regression probes, while the generative LLM results for earnings-surprise prediction are taken from \citet{fincall2026} and may involve different prompts and input processing. Third, our evaluation uses a single zero-shot prompt per task, and performance may be sensitive to prompt design. Fourth, both decision models are proprietary, and their performance may change with future model updates. Finally, our financial prediction experiments focus on earnings surprises and short-horizon stock movements using text-only inputs. These results may not generalize to other financial outcomes, longer prediction horizons, or settings combining text with numerical information.

\section*{Ethics Statement}
This study uses publicly available datasets and does not raise any significant ethical concerns.
\bibliography{custom}

@misc{typesafe2026jev,
  author       = {{TypeSafe AI}},
  title        = {Jev: A {System One} Decision Model},
  year         = {2026},
  howpublished = {\url{https://docs.typesafe.ai/concepts/system-one}},
  note         = {Model version \texttt{jev-1.13.0}. Accessed October 2026}
}

@misc{openai2026decisions,
  author       = {{OpenAI}},
  title        = {Decisions {API}},
  year         = {2026},
  howpublished = {\url{https://developers.openai.com/api/docs/guides/decisions}},
  note         = {Public beta, model \texttt{gpt-6-luna}. Accessed October 2026}
}

@misc{denney2026jev,
  author        = {Denney, Steven and DiGiuseppe, Matthew},
  title         = {{JEV} versus {LLMs}: Accuracy, Cost and Calibration on Seven Political Science Replications},
  year          = {2026},
  eprint        = {2610.06625},
  archivePrefix = {arXiv},
  primaryClass  = {cs.CL}
}

@article{brandt2025extractive,
  author  = {Brandt, Patrick T. and Alsarra, Sultan and D'Orazio, Vito and Heintze, Dagmar and Khan, Latifur and Meher, Shreyas and Osorio, Javier and Sianan, Marcus},
  title   = {Extractive versus Generative Language Models for Political Conflict Text Classification},
  journal = {Political Analysis},
  year    = {2025},
  pages   = {1--29},
  note    = {Published online ahead of print}
}

@article{huang2023finbert,
  author  = {Huang, Allen H. and Wang, Hui and Yang, Yi},
  title   = {{FinBERT}: A Large Language Model for Extracting Information from Financial Text},
  journal = {Contemporary Accounting Research},
  year    = {2023},
  volume  = {40},
  number  = {2},
  pages   = {806--841},
  doi     = {10.1111/1911-3846.12832}
}

@misc{yang2020finbert,
  author        = {Yang, Yi and Uy, Mark Christopher Siy and Huang, Allen},
  title         = {{FinBERT}: A Pretrained Language Model for Financial Communications},
  year          = {2020},
  eprint        = {2006.08097},
  archivePrefix = {arXiv},
  primaryClass  = {cs.CL}
}

@article{loughran2011liability,
  author  = {Loughran, Tim and McDonald, Bill},
  title   = {When Is a Liability Not a Liability? {Textual} Analysis, Dictionaries, and 10-{Ks}},
  journal = {The Journal of Finance},
  year    = {2011},
  volume  = {66},
  number  = {1},
  pages   = {35--65},
  doi     = {10.1111/j.1540-6261.2010.01625.x}
}

@inproceedings{li2023chatgpt,
  author    = {Li, Xianzhi and Chan, Samuel and Zhu, Xiaodan and Pei, Yulong and Ma, Zhiqiang and Liu, Xiaomo and Shah, Sameena},
  title     = {Are {ChatGPT} and {GPT-4} General-Purpose Solvers for Financial Text Analytics? {A} Study on Several Typical Tasks},
  booktitle = {Proceedings of the 2023 Conference on Empirical Methods in Natural Language Processing: Industry Track},
  year      = {2023},
  pages     = {408--422},
  address   = {Singapore},
  publisher = {Association for Computational Linguistics},
  doi       = {10.18653/v1/2023.emnlp-industry.39}
}

@inproceedings{guo2023chatgpt,
  author    = {Guo, Yue and Xu, Zian and Yang, Yi},
  title     = {Is {ChatGPT} a Financial Expert? {Evaluating} Language Models on Financial Natural Language Processing},
  booktitle = {Findings of the Association for Computational Linguistics: EMNLP 2023},
  year      = {2023},
  pages     = {815--821},
  address   = {Singapore},
  publisher = {Association for Computational Linguistics},
  doi       = {10.18653/v1/2023.findings-emnlp.58}
}

@inproceedings{xie2023pixiu,
  author    = {Xie, Qianqian and Han, Weiguang and Zhang, Xiao and Lai, Yanzhao and Peng, Min and Lopez-Lira, Alejandro and Huang, Jimin},
  title     = {{PIXIU}: A Comprehensive Benchmark, Instruction Dataset and Large Language Model for Finance},
  booktitle = {Advances in Neural Information Processing Systems 36 (NeurIPS 2023), Datasets and Benchmarks Track},
  year      = {2023}
}

@misc{wu2023bloomberggpt,
  author        = {Wu, Shijie and Irsoy, Ozan and Lu, Steven and Dabravolski, Vadim and Dredze, Mark and Gehrmann, Sebastian and Kambadur, Prabhanjan and Rosenberg, David and Mann, Gideon},
  title         = {{BloombergGPT}: A Large Language Model for Finance},
  year          = {2023},
  eprint        = {2303.17564},
  archivePrefix = {arXiv},
  primaryClass  = {cs.LG}
}

@inproceedings{tang2025finmteb,
  author    = {Tang, Yixuan and Yang, Yi},
  title     = {{FinMTEB}: Finance Massive Text Embedding Benchmark},
  booktitle = {Proceedings of the 2025 Conference on Empirical Methods in Natural Language Processing},
  year      = {2025},
  address   = {Suzhou, China},
  publisher = {Association for Computational Linguistics},
  doi       = {10.18653/v1/2025.emnlp-main.179}
}

@article{malo2014good,
  author  = {Malo, Pekka and Sinha, Ankur and Korhonen, Pekka and Wallenius, Jyrki and Takala, Pyry},
  title   = {Good Debt or Bad Debt: Detecting Semantic Orientations in Economic Texts},
  journal = {Journal of the Association for Information Science and Technology},
  year    = {2014},
  volume  = {65},
  number  = {4},
  pages   = {782--796},
  doi     = {10.1002/asi.23062}
}

@inproceedings{cortis2017semeval,
  author    = {Cortis, Keith and Freitas, Andr{\'e} and Daudert, Tobias and Huerlimann, Manuela and Zarrouk, Manel and Handschuh, Siegfried and Davis, Brian},
  title     = {{SemEval-2017} Task 5: Fine-Grained Sentiment Analysis on Financial Microblogs and News},
  booktitle = {Proceedings of the 11th International Workshop on Semantic Evaluation (SemEval-2017)},
  year      = {2017},
  pages     = {519--535},
  address   = {Vancouver, Canada},
  publisher = {Association for Computational Linguistics},
  doi       = {10.18653/v1/S17-2089}
}

@inproceedings{maia2018fiqa,
  author    = {Maia, Macedo and Handschuh, Siegfried and Freitas, Andr{\'e} and Davis, Brian and McDermott, Ross and Zarrouk, Manel and Balahur, Alexandra},
  title     = {{WWW'18} Open Challenge: Financial Opinion Mining and Question Answering},
  booktitle = {Companion Proceedings of the The Web Conference 2018},
  year      = {2018},
  pages     = {1941--1942},
  address   = {Lyon, France},
  publisher = {International World Wide Web Conferences Steering Committee},
  doi       = {10.1145/3184558.3192301}
}

@inproceedings{shah2023trillion,
  author    = {Shah, Agam and Paturi, Suvan and Chava, Sudheer},
  title     = {Trillion Dollar Words: A New Financial Dataset, Task \& Market Analysis},
  booktitle = {Proceedings of the 61st Annual Meeting of the Association for Computational Linguistics (Volume 1: Long Papers)},
  year      = {2023},
  pages     = {6664--6679},
  address   = {Toronto, Canada},
  publisher = {Association for Computational Linguistics},
  doi       = {10.18653/v1/2023.acl-long.368}
}

@misc{lu2023bbtfin,
  author        = {Lu, Dakuan and Wu, Hengkui and Liang, Jiaqing and Xu, Yipei and He, Qianyu and Geng, Yipeng and Han, Mengkun and Xin, Yingsi and Xiao, Yanghua},
  title         = {{BBT-Fin}: Comprehensive Construction of {Chinese} Financial Domain Pre-trained Language Model, Corpus and Benchmark},
  year          = {2023},
  eprint        = {2302.09432},
  archivePrefix = {arXiv},
  primaryClass  = {cs.CL}
}

@inproceedings{lan2023chinese,
  author    = {Lan, Yinyu and Wu, Yanru and Xu, Wang and Feng, Weiqiang and Zhang, Youhao},
  title     = {{Chinese} Fine-Grained Financial Sentiment Analysis with Large Language Models},
  booktitle = {FinLLM Symposium at IJCAI 2023},
  year      = {2023},
  eprint    = {2306.14096},
  archivePrefix = {arXiv}
}

@inproceedings{nan2021mdfend,
  author    = {Nan, Qiong and Cao, Juan and Zhu, Yongchun and Wang, Yanyan and Li, Jintao},
  title     = {{MDFEND}: Multi-Domain Fake News Detection},
  booktitle = {Proceedings of the 30th ACM International Conference on Information \& Knowledge Management (CIKM '21)},
  year      = {2021},
  pages     = {3343--3347},
  publisher = {Association for Computing Machinery},
  doi       = {10.1145/3459637.3482139}
}

@inproceedings{sinha2021impact,
  author    = {Sinha, Ankur and Khandait, Tanmay},
  title     = {Impact of News on the Commodity Market: Dataset and Results},
  booktitle = {Advances in Information and Communication (FICC 2021)},
  series    = {Advances in Intelligent Systems and Computing},
  volume    = {1364},
  pages     = {589--601},
  year      = {2021},
  publisher = {Springer},
  doi       = {10.1007/978-3-030-73103-8_41}
}

@inproceedings{sainz2023nlp,
  author    = {Sainz, Oscar and Campos, Jon and Garc{\'i}a-Ferrero, Iker and Etxaniz, Julen and Lopez de Lacalle, Oier and Agirre, Eneko},
  title     = {{NLP} Evaluation in Trouble: On the Need to Measure {LLM} Data Contamination for each Benchmark},
  booktitle = {Findings of the Association for Computational Linguistics: EMNLP 2023},
  year      = {2023},
  pages     = {10776--10787},
  address   = {Singapore},
  publisher = {Association for Computational Linguistics},
  doi       = {10.18653/v1/2023.findings-emnlp.722}
}

@inproceedings{xu2018stocknet,
  author    = {Xu, Yumo and Cohen, Shay B.},
  title     = {Stock Movement Prediction from Tweets and Historical Prices},
  booktitle = {Proceedings of the 56th Annual Meeting of the Association for Computational Linguistics (Volume 1: Long Papers)},
  year      = {2018},
  pages     = {1970--1979}
}

@inproceedings{wu2018hybrid,
  author    = {Wu, Huizhe and Zhang, Wei and Shen, Weiwei and Wang, Jun},
  title     = {Hybrid Deep Sequential Modeling for Social Text-Driven Stock Prediction},
  booktitle = {Proceedings of the 27th ACM International Conference on Information and Knowledge Management},
  year      = {2018},
  pages     = {1627--1630}
}

@inproceedings{soun2022accurate,
  author    = {Soun, Yejun and Yoo, Jaemin and Cho, Minyong and Jeon, Jihyeong and Kang, U},
  title     = {Accurate Stock Movement Prediction with Self-supervised Learning from Sparse Noisy Tweets},
  booktitle = {2022 IEEE International Conference on Big Data (Big Data)},
  year      = {2022},
  pages     = {1691--1700}
}

@inproceedings{fincall2026,
  author    = {Shu, Dong and Liu, Yanguang and Zhang, Huopu and Du, Mengnan},
  title     = {{FinCall-Surprise}: A Large Scale Multi-modal Benchmark for Earning Surprise Prediction},
  booktitle = {Proceedings of the 64th Annual Meeting of the Association for Computational Linguistics (Volume 1: Long Papers)},
  editor    = {Liakata, Maria and Moreira, Viviane P. and Zhang, Jiajun and Jurgens, David},
  month     = jul,
  year      = {2026},
  address   = {San Diego, California, United States},
  publisher = {Association for Computational Linguistics},
  pages     = {13357--13370},
  doi       = {10.18653/v1/2026.acl-long.610},
  url       = {https://aclanthology.org/2026.acl-long.610/}
}

\clearpage

\appendix


\onecolumn

\section{Prompts}
\label{app:prompts}

Every decision model request has one \emph{state} (the text to judge, in the fields listed) and one \texttt{choice} question with the instructions and option criteria below. Decision models received the structured instructions and criteria as JSON objects. 

\subsection{Classification tasks in English}

\begin{promptbox}{Financial PhraseBank}
\pfield{State fields}\texttt{sentence}\par
\pfield{Instructions}Using only the information explicitly stated in \texttt{sentence}, classify its financial sentiment from the viewpoint of an investor in the company mentioned. Choose the label representing the likely directional influence of the news on the relevant company's stock price. Do not use outside knowledge.\par
\pfield{Options}
\begin{itemize}
\item \texttt{negative}: The sentence contains economically unfavorable news likely to have a negative influence on the relevant company's stock price, such as deterioration, losses, declines, missed expectations, distress, or material risk.
\item \texttt{neutral}: The sentence is factual or descriptive and has no clear positive or negative economic influence on the relevant company's stock price, or opposing signals are balanced with no dominant direction.
\item \texttt{positive}: The sentence contains economically favorable news likely to have a positive influence on the relevant company's stock price, such as improvement, profits, growth, stronger demand, outperformance, or a beneficial development.
\end{itemize}
\end{promptbox}

\begin{promptbox}{FinSent}
\pfield{State fields}\texttt{sentence}\par
\pfield{Instructions}Using only \texttt{sentence}, classify the financial opinion or implication about the company or security discussed. Judge the dominant expressed sentiment, not isolated positive or negative words. Choose exactly one label.\par
\pfield{Options}
\begin{itemize}
\item \texttt{negative}: A clearly adverse financial evaluation or implication, such as deteriorating performance, losses, downside risk, distress, or unfavorable outlook.
\item \texttt{neutral}: Primarily factual, descriptive, balanced, or insufficiently directional about the focal company or security; no clearly dominant favorable or adverse implication.
\item \texttt{positive}: A clearly favorable financial evaluation or implication, such as improving performance, profit growth, upside potential, strong demand, or favorable outlook.
\end{itemize}
\end{promptbox}

\begin{promptbox}{SemEval-2017 Task 5 headlines}
\pfield{State fields}\texttt{sentence}\par
\pfield{Instructions}Using only \texttt{sentence}, classify the headline's dominant financial sentiment toward the company or market entity it discusses. Judge the economic implication rather than keyword tone. Choose exactly one label.\par
\pfield{Options}
\begin{itemize}
\item \texttt{negative}: The headline implies an unfavorable outcome or outlook for the focal entity, such as a loss, decline, warning, downgrade, or adverse development.
\item \texttt{neutral}: The headline is descriptive or mixed without a clearly positive or negative implication for the focal entity.
\item \texttt{positive}: The headline implies a favorable outcome or outlook for the focal entity, such as growth, stronger earnings, gains, upgrades, or beneficial developments.
\end{itemize}
\end{promptbox}

\begin{promptbox}{FiQA-ABSA}
\pfield{State fields}\texttt{sentence}\par
\pfield{Instructions}Using only \texttt{sentence}, classify the sentiment toward the explicitly named financial aspect. Ignore unrelated sentiment and choose exactly one label.\par
\pfield{Options}
\begin{itemize}
\item \texttt{negative}: The text expresses an unfavorable view of the named aspect, such as deterioration, risk, downgrade, loss, weakness, or disappointment.
\item \texttt{positive}: The text expresses a favorable view of the named aspect, such as improvement, growth, support, upgrade, gain, or optimism.
\end{itemize}
\end{promptbox}

\begin{promptbox}{Forward-looking statements (FLS)}
\pfield{State fields}\texttt{sentence}\par
\pfield{Question}Using only \texttt{sentence}, which forward-looking-statement category best describes its substantive meaning? Choose exactly one category.\par
\pfield{Method}Judge meaning and temporal orientation, not the mere presence of words such as will, expect, future, estimate, believe, may, or could.\par
\pfield{Options}
\begin{itemize}
\item \texttt{specific\_fls}: The sentence is about the future of the reporting company or another clearly identified company. It gives a company-specific expectation, plan, commitment, target, forecast, scheduled event, contractual future event, or expected company outcome. \emph{Not for:} Generic cautionary or hypothetical language that could apply to almost any company; general accounting rules; past or present facts.
\item \texttt{non\_specific\_fls}: The sentence is future-oriented, conditional, or hypothetical, but the statement is generic enough that it could be said of many companies. This includes boilerplate risks, cautionary language, generic possible consequences, and general future accounting or impairment conditions. \emph{Not for:} A concrete future plan, forecast, commitment, target, or event specific to the company; a sentence that is only about the past or present.
\item \texttt{not\_fls}: The sentence does not substantively discuss a future event or result. It reports historical or current facts, definitions, descriptions, or policies, including meta-language that merely describes forward-looking statements without itself making one. \emph{Not for:} An actual company-specific future statement or a generic future-oriented risk, condition, or consequence.
\end{itemize}
\end{promptbox}

\begin{promptbox}{ESG}
\pfield{State fields}\texttt{sentence}\par
\pfield{Question}Using only \texttt{sentence}, which ESG category best describes its primary substantive topic? Choose exactly one category.\par
\pfield{Method}Judge what the sentence is substantively about, not isolated keywords. If more than one topic appears, choose the category most central to the sentence's main claim. Use non\_esg when none of E, S, or G is substantive.\par
\pfield{Options}
\begin{itemize}
\item \texttt{environmental}: Effects on or management of the natural environment: climate change, greenhouse-gas emissions, energy and renewable energy, pollution, waste, water, biodiversity, ecosystems, natural resources, recycling, or environmental compliance and remediation. \emph{Not for:} Employee or community welfare without an environmental focus; board or ethical oversight; ordinary operational or financial facts.
\item \texttt{social}: Effects on or responsibilities toward people and stakeholders: employees, labor practices, workplace health and safety, diversity and inclusion, human rights, customers and product responsibility, suppliers, communities, education, philanthropy, access, or public health. \emph{Not for:} Natural-environment impacts alone; board structure, audit, corruption, or corporate-control matters; ordinary staffing or sales facts with no social responsibility content.
\item \texttt{governance}: How the company is directed, controlled, and held accountable: board and management oversight, shareholder rights, audit and internal controls, business ethics, codes of conduct, compliance systems, anti-corruption, lobbying and political accountability, executive compensation, or governance risk management. \emph{Not for:} General legal, accounting, or operational procedures without a corporate governance or ethical-accountability focus; employee or environmental topics.
\item \texttt{non\_esg}: Ordinary financial, accounting, commercial, operational, legal, or descriptive company information that is not substantively about environmental, social, or governance responsibility. \emph{Not for:} A sentence whose main claim concerns environmental impact, people or stakeholders, or corporate oversight and ethical accountability.
\end{itemize}
\end{promptbox}

\begin{promptbox}{FOMC hawkish--dovish}
\pfield{State fields}\texttt{sentence}\par
\pfield{Instructions}Using only \texttt{sentence}, classify the monetary-policy stance expressed in the sentence. Choose exactly one label.\par
\pfield{Options}
\begin{itemize}
\item \texttt{dovish}: The sentence supports easier or more accommodative monetary policy, lower rates, weaker economic conditions, or concern about downside risks.
\item \texttt{hawkish}: The sentence supports tighter or less accommodative monetary policy, higher rates, stronger inflation pressure, or concern about overheating.
\item \texttt{neutral}: The sentence describes facts, methods, background, or economic conditions without a clear directional implication for tighter or easier monetary policy.
\end{itemize}
\end{promptbox}

FinancialFraud filings longer than 16{,}000 characters were truncated to a deterministic excerpt: the first 3{,}000 characters, a 1{,}000-character window around the first match of each fraud-related keyword, and the last 2{,}000 characters.

\begin{promptbox}{FinancialFraud}
\pfield{State fields}\texttt{sentence}\par
\pfield{Instructions}Using only \texttt{sentence}, decide whether the excerpt itself provides clear affirmative evidence that its source filing is associated with financial fraud. Choose exactly one label; do not equate ordinary losses, risks, or accounting terminology with proven fraud.\par
\pfield{Options}
\begin{itemize}
\item \texttt{no}: The excerpt does not provide clear evidence of intentional financial misrepresentation or a fraud finding; routine accounting, risks, or poor performance alone are not fraud.
\item \texttt{yes}: The excerpt contains clear evidence of intentional financial misstatement, deception, fabrication, or an explicit fraud finding tied to the source filing.
\end{itemize}
\end{promptbox}

\subsection{Classification tasks in Chinese}

\begin{promptbox}{FinNSP}
\pfield{State fields}\texttt{sentence}\par
\pfield{Instructions}\zh{只根据 \texttt{sentence}，判断文本是否包含明确针对某个金融机构、公司、平台或其他金融主体的负面消息。此处只做是否存在负面消息的二分类，不要求抽取主体。选择一个类别。}\par
\pfield{Options}
\begin{itemize}
\item \texttt{non\_negative\_news}: \zh{没有可明确归属于金融主体的负面事件或指控；普通市场数据、客观介绍、无关广告、一般风险提示或负面词汇本身不足以算负面消息。}
\item \texttt{negative\_news}: \zh{明确报道或指控金融主体遭遇或实施不利事件，例如诈骗、违规、违约、跑路、重大亏损、处罚、失信或严重经营问题。}
\end{itemize}
\end{promptbox}

\begin{promptbox}{FinChinaSentiment}
\pfield{State fields}\texttt{sentence}\par
\pfield{Instructions}\zh{只根据 \texttt{sentence}，判断报道对其中主要企业或机构的负面预警强度。数据集以企业预警为目的：0 表示非负面，-1、-2、-3 表示逐级加重的负面。若报道涉及多个主体且没有明确目标，以标题和主体事件为主。选择一个类别。}\par
\pfield{Options}
\begin{itemize}
\item \texttt{0}: \zh{未形成明确的企业负面预警；以中性事实、正面进展或无法指向具体主体的波动为主。}
\item \texttt{-1}: \zh{一般负面：对企业或机构有明确不利影响，但影响较轻、范围有限、仍属常规经营或争议风险。}
\item \texttt{-2}: \zh{很消极：较严重的亏损、重大经营困难、监管调查、重大安全或合规事件等，明显威胁企业经营或声誉。}
\item \texttt{-3}: \zh{特别消极：极严重的违法犯罪、重大欺诈、重罚、严重危机或足以造成重大持续损害的事件。}
\end{itemize}
\end{promptbox}

\begin{promptbox}{FinFE}
\pfield{State fields}\texttt{sentence}\par
\pfield{Instructions}\zh{只根据 \texttt{sentence}，判断金融社交媒体文本对所讨论股票、公司或市场的主要情绪方向。识别反讽、股民口语和涨跌含义；不要仅凭一个情绪词判断。选择一个类别。}\par
\pfield{Options}
\begin{itemize}
\item \texttt{negative}: \zh{明确看空、失望、批评、担忧亏损或下跌，或表达明显不利的市场预期。}
\item \texttt{neutral}: \zh{讨论、提问、单纯报数或方向不明；没有明显看多或看空态度。}
\item \texttt{positive}: \zh{明确看多、乐观、称赞、预期上涨、盈利或利好。}
\end{itemize}
\end{promptbox}

\begin{promptbox}{OpenFinDataSentiment}
\pfield{State fields}\texttt{sentence}\par
\pfield{Instructions}\zh{只根据 \texttt{sentence}，判断这段中文金融文本的整体情绪或对主要市场主体的经济含义。选择一个类别。}\par
\pfield{Options}
\begin{itemize}
\item \zh{\texttt{消极}：主要表达负面评价，或描述亏损、下跌、恶化、风险等明显不利结果。}
\item \zh{\texttt{中性}：主要是客观陈述、信息罗列或正反信号相抵，没有明显主导情绪。}
\item \zh{\texttt{积极}：主要表达正面评价，或描述增长、改善、盈利、上涨等明显有利结果。}
\end{itemize}
\end{promptbox}

\begin{promptbox}{MDFEND-Weibo21}
\pfield{State fields}\texttt{sentence}\par
\pfield{Instructions}\zh{只根据 \texttt{sentence}，判断这条财经微博是否更像已核实的真实消息或谣言/虚假消息。不要把负面、耸动或未附链接本身当成虚假；缺乏外部证据时只能作文本线索判断。选择一个类别。}\par
\pfield{Options}
\begin{itemize}
\item \texttt{real}: \zh{没有明显的编造或已辟谣线索，叙述相对具体、一致；这不等于已经被独立核实。}
\item \texttt{fake}: \zh{文本内部呈现明显的谣言、伪造、被辟谣、严重自相矛盾或不实传播线索。}
\end{itemize}
\end{promptbox}

\subsection{Headline and question pair tasks}

\begin{promptbox}{HeadlinePDU}
\pfield{State fields}\texttt{text\_a}, \texttt{text\_b}\par
\pfield{Question}Using only \texttt{text\_a} and \texttt{text\_b}, decide whether the two financial headlines match on the price-up dimension. Choose exactly one label.\par
\pfield{Options}
\begin{itemize}
\item \texttt{not\_match}: The headlines do not express the same upward-price or bullish financial dimension, even if they share an asset or a few words.
\item \texttt{match}: Both headlines express a materially similar price-up, gain, bullish, or favorable market-movement dimension for the relevant asset or market.
\end{itemize}
\end{promptbox}

\begin{promptbox}{HeadlinePDD}
\pfield{State fields}\texttt{text\_a}, \texttt{text\_b}\par
\pfield{Question}Using only \texttt{text\_a} and \texttt{text\_b}, decide whether the two financial headlines match on the price-down dimension. Choose exactly one label.\par
\pfield{Options}
\begin{itemize}
\item \texttt{not\_match}: The headlines do not express the same downward-price or bearish financial dimension, even if they share an asset or a few words.
\item \texttt{match}: Both headlines express a materially similar price-down, loss, decline, bearish, or unfavorable market-movement dimension for the relevant asset or market.
\end{itemize}
\end{promptbox}

\begin{promptbox}{HeadlineAC}
\pfield{State fields}\texttt{text\_a}, \texttt{text\_b}\par
\pfield{Question}Using only \texttt{text\_a} and \texttt{text\_b}, decide whether the two financial headlines match on the asset-comparison dimension. Choose exactly one label.\par
\pfield{Options}
\begin{itemize}
\item \texttt{not\_match}: The headlines do not express the same asset-comparison relationship or comparable asset-market dimension. Shared generic finance words alone are insufficient.
\item \texttt{match}: Both headlines describe a materially similar relationship among the same or comparable assets, commodities, markets, or relative asset movements. The two headlines may have opposite price directions.
\end{itemize}
\end{promptbox}

\begin{promptbox}{AFQMC}
\pfield{State fields}\texttt{text\_a}, \texttt{text\_b}\par
\pfield{Question}Using only \texttt{text\_a} and \texttt{text\_b}, decide whether the two Chinese financial questions express the same user intent. Choose exactly one label.\par
\pfield{Options}
\begin{itemize}
\item \texttt{not\_match}: The questions ask for different information, actions, objects, or conditions, even if they mention the same financial product or entity.
\item \texttt{match}: The questions are semantically equivalent or ask for the same answer about the same financial product, account, transaction, or condition, allowing natural paraphrase and minor wording differences.
\end{itemize}
\end{promptbox}

\subsection{Prediction tasks}

For FinCall-Surprise, the state is the transcript. Following \citet{fincall2026}, transcripts longer than 31K tokens (\texttt{cl100k\_base}) were shortened for \jev{} and \luna{} by summarising speaker turns with BART-Large-CNN, Operator turns first, then Executives, then Analysts, until the transcript fit.

\begin{promptbox}{FinCall-Surprise}
\pfield{State fields}\texttt{transcript}\par
\pfield{Instructions}You are an expert equity research analyst. Based only on this earnings conference-call transcript, predict whether the company's next-quarter EPS will beat or miss consensus estimates. Use the transcript's forward-looking signals, operating outlook, guidance, demand, costs, and risks.\par
\pfield{Options}
\begin{itemize}
\item \texttt{positive}: Expected positive earnings surprise: indications of stronger-than-expected EPS or revenue growth, upbeat guidance, successful launches or partnerships, cost savings, or expansion.
\item \texttt{negative}: Expected negative earnings surprise: indications of missed earnings or revenue declines, weak or withdrawn guidance, regulatory or legal setbacks, unexpected costs, competitive pressure, or deteriorating outlook.
\end{itemize}
\end{promptbox}

For stock movement, the state lists the tweets about the stock from the five calendar days before the target day, with at most 30 tweets of at most 40 tokens per trading day. The state is defined as:

\begin{quote}\small\ttfamily
Stock: \$\textit{TICKER}\\[2pt]
Tweets before trading day \textit{YYYY-MM-DD}:\\
- \textit{tweet}\\
\dots\\
Tweets before target day:\\
- \textit{tweet}
\end{quote}

\begin{promptbox}{StockNet, CIKM18 and BigData22}
\pfield{State fields}\texttt{tweets}\par
\pfield{Instructions}You are an equity trader. Below are tweets about one stock from the five days before a trading day, grouped by the trading day they precede. Based only on these tweets, predict whether the stock's adjusted closing price will rise or fall on the target trading day relative to the previous close.\par
\pfield{Options}
\begin{itemize}
\item \texttt{up}: The stock closes higher on the target day: the tweets suggest buying interest, positive news, upgrades, strong results, or bullish momentum.
\item \texttt{down}: The stock closes lower on the target day: the tweets suggest selling pressure, negative news, downgrades, weak results, or bearish momentum.
\end{itemize}
\end{promptbox}

\end{document}